\documentclass[10pt,twocolumn,letterpaper]{article}

\usepackage{setup/wacv/wacv}              
\usepackage[accsupp]{axessibility}  

\usepackage{epsfig}
\usepackage{graphicx}
\usepackage{subcaption}
\usepackage{float}
\usepackage[justification=justified]{caption}	    
\usepackage{lscape}                                         
\usepackage{wrapfig}
\usepackage[inkscapeformat=png]{svg}

\usepackage[lined,ruled,linesnumbered]{algorithm2e}
\usepackage{animate} 

\usepackage{booktabs}                   
\usepackage{multirow}

\usepackage{makecell}

\usepackage{paralist}
\usepackage{enumitem}

\usepackage{tabularray}                 
\UseTblrLibrary{booktabs}               

\usepackage{bm}                          
\usepackage{bbm}                         
\usepackage{epsfig}                      
\usepackage{graphicx}                  
\usepackage{times}
\usepackage{mathtools}
\usepackage{amssymb,amsmath}   

\usepackage{units}
\usepackage{color}
\usepackage[T1]{fontenc}    
\usepackage{amsfonts}       
\usepackage[utf8]{inputenc} 
\usepackage{nicefrac}       
\usepackage{microtype}      
\usepackage[normalem]{ulem} 
\usepackage{comment}

\usepackage{url}  
\usepackage{xspace}
\usepackage{xcolor}

\def\eg{e.g.,~}               
\def\ie{i.e.,~}               
\def\vs{vs.~}                 

\DeclareMathOperator*{\argmin}{\arg\!\min}
\DeclareMathOperator*{\argmax}{\arg\!\max}

\newlength\paramargin
\newlength\figmargin

\newlength\secmargin
\newlength\figcapmargin
\newlength\tabcapmargin

\newcommand{\mpage}[2]
{
\begin{minipage}{#1\linewidth}\centering
#2
\end{minipage}
}

\newcommand{\figcaption}[2]
{
\caption{
\textbf{#1.}  
#2            
}
}

\newcommand{\secref}[1]{Section~\ref{sec:#1}}
\newcommand{\figref}[1]{Figure~\ref{fig:#1}}
\newcommand{\tabref}[1]{Table~\ref{tab:#1}}
\newcommand{\eqnref}[1]{\eqref{eq:#1}}

\long\def\ignorethis#1{}

\newbox\jsavebox%

\newcommand{\best}[1]{{\textbf{#1}}}
\newcommand{\second}[1]{{\underline{#1}}}

\def\xi{x_i}

\def\noteg{^{g}}
\def\notel{^{l}}
\def\notex{^\text{cross}}
\def\adv{\text{adv}}

\def\cop{\text{TOL}}

\newcommand{\kb}{{Kinetics$\rightarrow$BABEL}}

\graphicspath{{figure}, {example}}

\usepackage[pagebackref,breaklinks,colorlinks]{hyperref}

\usepackage[capitalize]{cleveref}
\crefname{section}{Sec.}{Secs.}
\Crefname{section}{Section}{Sections}
\Crefname{table}{Table}{Tables}
\crefname{table}{Tab.}{Tabs.}

\def\wacvPaperID{1528} 
\def\confName{WACV}
\def\confYear{2024}

\begin{document}

\title{GLAD: Global-Local View Alignment and Background Debiasing\\
for Unsupervised Video Domain Adaptation with Large Domain Gap}


\author{
Hyogun Lee\textsuperscript{1}\thanks{Equal contributor} ,
Kyungho Bae\textsuperscript{1}\footnotemark[1] ,
Seong Jong Ha$^2$,
Yumin Ko$^3$, \\
Gyeong-Moon Park\textsuperscript{1}\thanks{Corresponding author} ,
Jinwoo Choi\textsuperscript{1}\footnotemark[2] , \\
$^1$Kyung Hee University, $^2$AI Center, CJ Corporation, $^3$NCSOFT \\
{\tt\small \{gunsbrother,kyungho.bae,gmpark,jinwoochoi\}@khu.ac.kr} \\
{\tt\small oanchovy@cj.net,yuminko@ncsoft.com}
}

\twocolumn[{
\renewcommand\twocolumn[1][]{#1}
\maketitle

\begin{center}
    \centering
    \includegraphics[width=\linewidth]{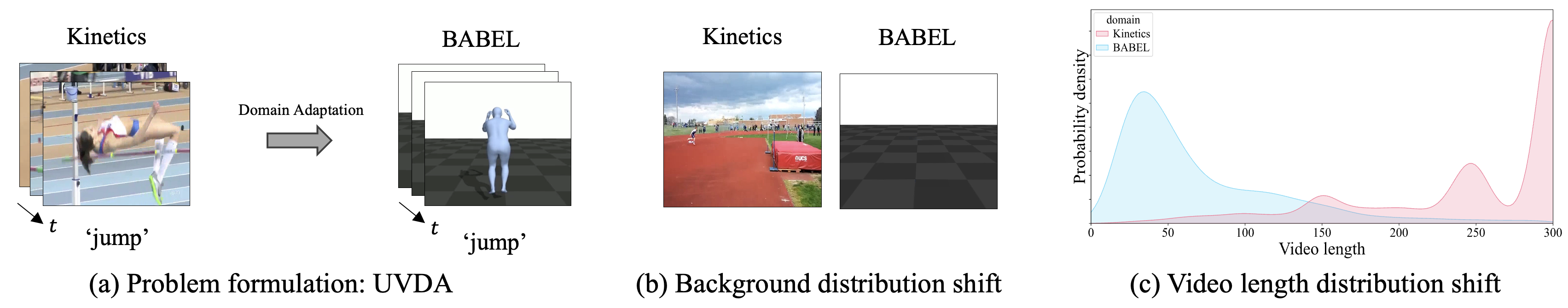}
    
    \captionof{figure}{\textbf{Overview of {\kb} dataset.}
    We introduce a challenging unsupervised domain adaptation (UDA) dataset, \kb. 
    (a) We formulate the problem of action recognition as UDA where we have labeled source dataset, \eg Kinetics, and unlabeled target dataset, \eg BABEL. The dataset presents two challenges: (b) Background distribution shift: The source dataset (Kinetics) exhibits diverse backgrounds, while the target dataset (BABEL) consistently features the same background across videos. (c) Video length distribution shift: Videos in the source dataset (Kinetics) tend to be longer, while videos in the target dataset (BABEL) are typically shorter. These challenges make the {\kb} dataset a valuable benchmark for studying UDA for action recognition.
    }
    
    \label{fig:teaser}
\end{center}

}]

\renewcommand{\thefootnote}{\fnsymbol{footnote}}
\footnotetext[1]{Equally contributed first authors.}
\footnotetext[2]{Corresponding authors.}


\begin{abstract}

In this work, we tackle the challenging problem of unsupervised video domain adaptation (UVDA) for action recognition. We specifically focus on scenarios with a substantial domain gap, in contrast to existing works primarily deal with small domain gaps between labeled source domains and unlabeled target domains. To establish a more realistic setting, we introduce a novel UVDA scenario, denoted as \kb, with a more considerable domain gap in terms of both temporal dynamics and background shifts. To tackle the temporal shift, \ie action duration difference between the source and target domains, we propose a global-local view alignment approach. To mitigate the background shift, we propose to learn temporal order sensitive representations by temporal order learning and background invariant representations by background augmentation. We empirically validate that the proposed method shows significant improvement over the existing methods on the {\kb} dataset with a large domain gap. 
The code is available at \url{https://github.com/KHU-VLL/GLAD}.

\end{abstract}

\section{Introduction}
\label{sec:intro}










Human action recognition in videos is an interesting problem in computer vision. There are immense practical applications of action recognition: video surveillance, retrieval, captioning, sports analysis, health care, and autonomous driving. Achieving accurate and robust action recognition performance enables improved security and efficient video analysis.

Recent advances in action recognition have witnessed remarkable progress, primarily attributed to the availability of extensive labeled datasets and the successful deployment of deep learning architectures, such as convolutional neural networks (CNNs)~\cite{carreira2018i3d,feichtenhofer2019slowfast,lin2019tsm,tran2015c3d} and transformers~\cite{arnab2021vivit,bertasius2021timesformer,liu2022video,patrick2021motionformer}. However, collecting large-scale annotated video data remains a challenging and costly endeavor due to the additional temporal dimension compared to image annotation. Due to the high annotation cost, labeled video datasets do not scale sufficiently, resulting in poor generalization in unseen domain~\cite{choi2019cant}. 

To address the aforementioned challenge of poor generalization, an effective approach is to formulate the action recognition task as an unsupervised domain adaptation (UDA) problem. In the UDA setting, we leverage a labeled source dataset to achieve good performance on an \emph{unlabeled} target dataset. The recent works on unsupervised video domain adaptation (UVDA) for action recognitionhave shown impressive performance improvement~\cite{chen2019ta3n,choi2020sava,da2022co2a,munro2020mm-sada,pan2019tcon,sahoo2021comix,song2021STCDA,yang2022cia}  on the standard UCF-HMDB~\cite{chen2019ta3n} and EPIC-KITCHENS~\cite{munro2020mm-sada} datasets.

However, the impressive performance on the UCF-HMDB and EPIC-KITCHENS datasets may not necessarily reflect real-world scenarios. This discrepancy arises due to several reasons. Firstly, these datasets have a relatively small scale. The UCF-HMDB dataset consists of 3,209 videos from both the source and target domains, which is considerably smaller compared to the original UCF-101~\cite{soomro2012ucf101} and HMDB-51~\cite{kuehne2011hmdb} datasets. This limited data can lead to overfitting issues as models struggle to effectively generalize. Secondly, the UCF-HMDB and EPIC-KITCHENS datasets do not exhibit significant domain gaps.
As shown in \tabref{dataset-comapare}, 
the accuracy gap between the model trained with target labels and the model trained with only the source data and labels is 11.4 points for UCF-HMDB and 26.2 points for EPIC-KITCHENS. 
However, real-world scenarios often involve more substantial domain gaps, such as the real-synthetic gap, day-night gap, sunny-snowy gap, and others. These domain gaps present additional challenges that need to be addressed for action recognition models to reliably perform in diverse and complex environments. 

\begin{figure}[t]
\centering
    \includegraphics[page=1,width=\linewidth]{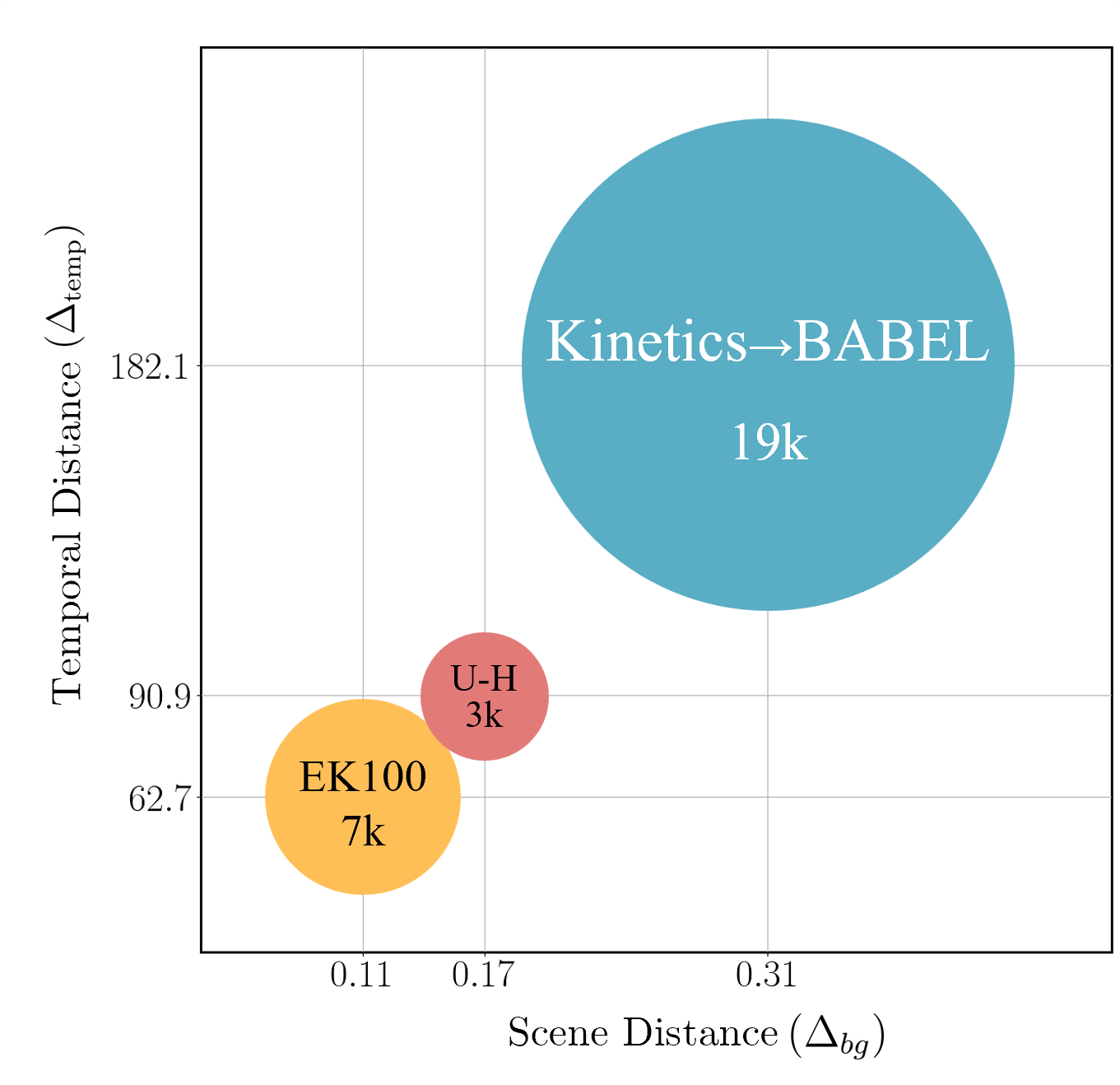}
    \vspace{\figcapmargin}

    \figcaption{Comparison between Kinetics$\rightarrow$BABEL and the existing UVDA datasets}{
       We compare the scene distance ($\Delta_{bg}$), the temporal distance ($\Delta_\text{temp}$), and the scale of the UCF-HMDB, EPIC-KITCHENS, and the proposed Kinetics$\rightarrow$BABEL datasets. The Kinetics$\rightarrow$BABEL dataset shows more substantial domain gaps between the source and target, and is much larger than the existing datasets.
       }

    \label{fig:K_b_dist}
\end{figure}

To address the limitations of existing datasets, we introduce Kinetics$\rightarrow$BABEL, a new and comprehensive dataset designed to present greater challenges for unsupervised video domain adaptation. The {\kb} dataset significantly expands the scale, comprising a total of 18,946 videos. As depicted in \figref{teaser}, the {\kb} dataset exhibits substantial temporal and background distribution shifts between the source and target domains. In \figref{teaser} (c), it is evident that the videos from the Kinetics dataset tend to be longer compared to the videos from BABEL. Furthermore, the background distributions differ between the two datasets, with Kinetics displaying real but biased backgrounds for different actions, while BABEL features a consistent gray-scale checkerboard background across actions as shown in \figref{teaser} (b). 
In \figref{K_b_dist}, we compare the proposed Kinetics$\rightarrow$BABEL dataset with existing datasets in terms of the scene distance ($\Delta_{bg}$), temporal distance ($\Delta_\text{temp}$), and scale. The Kinetics$\rightarrow$BABEL dataset shows more substantial domain gaps between the source and target, and is much larger than the existing datasets. The proposed dataset is much more realistic and challenging compared to the existing datasets. Please refer to \secref{kinetics->babel} for more details on the dataset.



To tackle the challenging UVDA with a large domain gap in \kb, we propose i) \textbf{G}lobal-\textbf{L}ocal view \textbf{A}lignment and ii) background \textbf{D}ebiasing for unsupervised video domain adaptation (\textbf{GLAD}).
i) To address the temporal duration shift between the source and target domains, we propose a Global-Local temporal view Alignment approach, GLA. GLA aligns a set of source clips, sampled at diverse temporal sampling rates, with a set of target clips that also exhibit varying sampling rates. By considering global and local temporal perspectives, our approach facilitates the learning of domain-invariant representations, particularly effective in scenarios with large temporal shifts.
ii) To address the background distribution shift between the source and target domains, we propose a background-invariant representation learning to debias background bias, inspired by prior works~\cite{choi2020sava,sahoo2021comix}. The proposed debiasing method leverages both background augmentation via background mixing and temporal order learning. By incorporating these techniques, we mitigate the impact of background distribution shift between domains, thereby improving the performance on the target domain. To validate the efficacy of our proposed method, we conduct extensive empirical evaluations on the challenging {\kb} dataset. Our experimental results demonstrate the superiority of GLAD in handling UVDA with a significant domain gap, showcasing its effectiveness in achieving robust action recognition performance in real-world scenarios. To facilitate further research, we plan to publicly release the {\kb} dataset and code upon acceptance of this paper.

In summary, our work makes the following key contributions:
\begin{itemize}
    \item We introduce the novel {\kb} dataset, specifically designed for unsupervised video domain adaptation with a substantial domain gap. The {\kb} dataset exhibits significantly larger temporal and background distribution shifts compared to existing datasets, making it a more challenging and realistic benchmark.

    \item To tackle the temporal and background shifts between the source and target domains, we propose a novel approach called Global-Local view Alignment and background Debiasing (GLAD). GLAD incorporates global-local view alignment techniques to address temporal shifts and employs background debiasing methods to mitigate the background distribution shift.
    
    \item We empirically demonstrate the effectiveness of the proposed method via extensive experiments on the challenging {\kb} dataset.
\end{itemize}

\section{Related Work}
\label{sec:related-work}
\paragraph{Action Recognition.}
Deep neural networks have demonstrated remarkable progress in the field of action recognition. Various approaches have been explored to recognize actions from videos. One common approach is to utilize 2D CNNs~\cite{donahue2016longterm,karpathy20142dcnn,lin2019tsm,Simonyan-NIPS-2014,ng2015short,zhou2018trn}, which extract features from individual frames of a video and incorporate temporal modeling techniques. Another popular approach involves 3D CNNs~\cite{carreira2018i3d,feichtenhofer2019slowfast,ji20123d,tran2015c3d}, which learn to capture spatio-temporal features from short video clips. Recently, transformers with spatio-temporal attention mechanisms have also demonstrated impressive performance in action recognition~\cite{arnab2021vivit,bertasius2021timesformer,patrick2021motionformer}. However, most of the existing action recognition methods heavily rely on large amounts of labeled data. In contrast, our work takes a different approach by formulating the action recognition problem as unsupervised domain adaptation. In this setting, we no longer require labeled data from the target domain, but instead leverage labeled data from the source domain.

\paragraph{Unsupervised Domain Adaptation.}
In recent years, substantial efforts have been dedicated to unsupervised domain adaptation (UDA) for both image domains~\cite{dann,saito2018maximum,zhang2021image_DA_survey,zhang2019domain} and video domains (UVDA)~\cite{chen2019ta3n,choi2020sava,kim2021CMCo,sahoo2021comix,wei2022transvae,xu2022ATCoN,yang2022cia}.
To tackle the UVDA problem, adversarial-based methods \cite{chen2019ta3n,munro2020mm-sada,pan2019tcon}, semantic-based methods \cite{da2022co2a,sahoo2021comix}, and self-supervised methods~\cite{choi2020sava,munro2020mm-sada} have shown significant progress.
However, the majority of existing UVDA works evaluate their performance on small-scale and less challenging datasets such as UCF-HMDB~\cite{chen2019ta3n} or EPIC-KITCHENS~\cite{munro2020mm-sada}.
This limitation hampers the comprehensive evaluation of UVDA methods in more demanding scenarios. To address this gap, we introduce a novel and large-scale UVDA dataset called \kb, which exhibits a significant domain gap. Our proposed method is specifically designed to tackle the challenges presented by this dataset. We anticipate that the {\kb} dataset serves as a new standard benchmark for evaluating UVDA methods, facilitating further advancements in this field.

\paragraph{Background bias.}
The research community has recognized background bias as a significant challenge in video action recognition~\cite{choi2019cant,li2018resound,li2019repair}. When an action recognition model is biased toward the background, it relies on spurious correlations between actions and backgrounds rather than understanding the true semantics of the human actions. The background bias becomes even more detrimental in the context of UVDA, where the model needs to adapt to a target domain with different background distributions without action labels. Several approaches demonstrate the benefits of background debiasing in UVDA~\cite{choi2020sava,sahoo2021comix}. In this work, we also address the significant background bias present in the source domain, Kinetics, aiming to achieve favorable performance on the target domain, BABEL, which exhibits entirely different background distributions. By mitigating the background bias, we encourage the action recognition model to focus on genuine action semantics and enhance its ability to adapt to diverse target domains with varying background characteristics.

\newcommand{\norm}[1]{\left\lVert#1\right\rVert}
\newcommand{\domS}{$ \mathcal{D}_\text{source} $}
\newcommand{\domT}{$ \mathcal{D}_\text{target} $}
\newcommand{\ui}{\textbf{u}_i}
\newcommand{\vj}{\textbf{v}_j}
\newcommand{\E}{\mathbb{E}}

\section{Kinetics$\rightarrow$BABEL Dataset}
\label{sec:kinetics->babel}

We introduce a new dataset called {\kb}, designed to evaluate the performance of UVDA methods in a more realistic and challenging setting. In this work, we set Kinetics as the source domain and BABEL as the target domain. The {{\kb}} dataset is constructed by re-organizing two existing datasets: Kinetics~\cite{kay2017kinetics} and BABEL~\cite{punnakkal2021babel}. Kinetics$\rightarrow$BABEL consists of 12 classes, specifically selected from the overlapping classes of Kinetics and BABEL: {\small {\texttt{jump, run, throw, kick, bend, dance, clean something, squat, punch, crawl, clap, pick up}}}. The dataset comprises 14,881 training and 650 test videos from the Kinetics dataset, and 2,963 training and 452 test videos from the BABEL dataset.

The proposed UVDA dataset encompasses both the real-world Kinetics dataset and the synthetic BABEL dataset. Leveraging synthetic datasets is cost-effective compared to real-world data collection, making their integration as source or target datasets a commonly adopted approach. As shown in the previous works~\cite{hoffman2018cycada,tsai2018learning,chen2019ta3n}, real-to-synthetic and synthetic-to-real domain adaptation problems are quite challenging which makes the proposed dataset interesting. In this work, we focus on the Kinetics$\rightarrow$BABEL domain adaptation setting, leaving BABEL$\rightarrow$Kinetics domain adaptation setting as a future work.

The Kinetics$\rightarrow$BABEL domain adaptation presents two significant challenges: the appearance gap and the temporal gap between the source and target data. The BABEL dataset lacks background information, in contrast to the Kinetics dataset which consists of videos with realistic backgrounds. Moreover, while Kinetics videos exhibit similar durations, BABEL videos encompass a wider range of durations. Consequently, addressing both the background and temporal gaps in a comprehensive domain adaptation strategy becomes crucial to achieve a good performance on the Kinetics$\rightarrow$BABEL dataset.

Notably, the proposed {\kb} dataset exhibits a larger domain gap compared to existing UVDA datasets, such as UCF-HMDB~\cite{chen2019ta3n} and EPIC-KITCHENS~\cite{munro2020mm-sada}. To quantify the background gap, denoted as $\Delta_{bg}$, we calculate the average minimum scene feature distance between each source video and all target videos and vice versa as follows:
\begin{align}
\Delta_{bg} &= \frac{1}{2}[\frac{1}{L_S}\sum_{i=1}^{L_S} \min_j d(\mathbf{u}_i, \mathbf{v}_j) + \frac{1}{L_T} \sum_{j=1}^{L_T} \min_i d(\mathbf{u}_i, \mathbf{v}_j)].
\label{eq:dist_scene}
\end{align}
Here, $\mathbf{u}_i$ represents the scene feature vector of the source domain with $L_S$ videos, $\mathbf{v}_j$ denotes the scene feature vector of the target domain with $L_T$ videos, and $d(\mathbf{u}, \mathbf{v}) = 1 - \mathbf{u}^\mathrm{T}\mathbf{v}$ is the cosine distance between them. We employ a ResNet-50~\cite{he2016resnet} model pre-trained on the Places365 dataset~\cite{zhou2017places} to extract scene features.

Furthermore, {\kb} also shows the huge domain gap in the temporal perspective. To assess the temporal gap, we leverage the earth mover's distance (EMD)~\cite{rubner2000earth, kantorovich1942translocation}. The EMD quantifies the minimal cost required to transform one distribution into another, providing an intuitive measure of similarity between distributions. 
We compute the EMD between two video length distributions $p, q$ as follows:
\begin{equation}
    \Delta_\text{temp} = \text{EMD}(p, q) = \int |\text{CDF}_p(x) - \text{CDF}_q(x)| \, dx \, .
\label{eq:dist_temp}
\end{equation}

In \tabref{dataset-comapare}, we show three domain gaps between the source and target data: the scene distance ($\Delta_{bg}$), the temporal distance ($\Delta_\text{temp}$), and the accuracy gap ($\Delta_\text{Acc}$) for various UVDA datasets. It is evident that both the UCF-HMDB and EPIC-KITCHENS datasets exhibit relatively smaller scene distances of 0.17 and 0.11 respectively. In contrast, the proposed {\kb} dataset demonstrates a significantly larger scene distance of 0.31, indicating a more pronounced background gap between the domains.
Furthermore, {\kb} shows a more realistic temporal gap for UVDA settings. The temporal distance of Kinetics$\rightarrow$BABEL is 182.1 frames which is $2 \times$ bigger than the temporal gap of the UCF-HMDB and $3 \times$ bigger than the temporal gap of the EPIC-KITCHENS. To achieve good performance on the Kinetics$\rightarrow$BABEL dataset, a model should be able to focus on the action instead of the background as well as learn to represent videos with various lengths.

Due to the presence of background and temporal gaps, the model performance decreases on the target domain, as indicated by $\Delta_\text{Acc}$. $\Delta_\text{Acc}$ denotes the performance gap between the model trained with target labels and the model trained with the source data and labels only. The {\kb} shows a significant gap of 65.0 points while UCF-HMDB shows 11.4 points and EPIC-KITCHENS shows 26.2 points respectively. Moreover, the {\kb} dataset comprises 18,946 videos, making it substantially larger in scale compared to both UCF-HMDB (3,209 videos) and EPIC-KITCHENS (6,729 videos). These observations clearly demonstrate that the proposed {\kb} dataset is a large-scale and challenging benchmark to properly evaluate the performance of unsupervised video domain adaptation methods.


 \paragraph{Comparison with other synthetic-real datasets.}
There are a few synthetic-real datasets for the problem of domain-adaptive action recognition: Kinetics-Gameplay~\cite{chen2019ta3n} and Mixamo-Kinetics~\cite{da2022co2a}. Compared to these existing datasets, the proposed dataset has some advantages. As shown in \tabref{dataset-comapare}, the proposed Kinetics$\rightarrow$BABEL dataset offers the larger scene distance ($\Delta_{bg}$) and temporal distance ($\Delta_{temp}$) between domains, compared to the existing synthetic-real datasets. Also, note that the raw RGB data of the Kinetics-Gameplay dataset is not publicly available while we make the raw data of the Kinetics$\rightarrow$BABEL dataset public.

\begin{table}[t]
\centering 

\caption{
\textbf{UVDA dataset statistics.}
We provide a quantitative evaluation of commonly used benchmarks in the field of UVDA. The table includes the number of shared classes (\# classes), the total number of videos (\# videos), 
the scene distance ($\Delta_{bg}$) in frames calculated by \eqnref{dist_scene} and the temporal distance ($\Delta_\text{temp}$) in frames calculated by \eqnref{dist_temp}, and the accuracy gap ($\Delta_\text{Acc}$) between ``target only'' and ``source only'' performances. The best quantities are in bold.
}


\resizebox{\columnwidth}{!}{
\begin{tblr}{ lrrcc }
\toprule
Dataset& \# classes & \# videos & $\Delta_{bg}$ & $\Delta_\text{temp}$ & \hphantom{\textsuperscript{\textdagger}}$\Delta_\text{Acc}$   \\
\midrule
UCF-HMDB \cite{chen2019ta3n} & \second{12} & 3,209 & 0.17 & \second{90.9} & \hphantom{\textsuperscript{\textdagger}}11.4 \\ 
EPIC-KITCHENS UDA \cite{munro2020mm-sada} & 8 & $^*$6,729 & 0.11 & 62.7 & \hphantom{\textsuperscript{\textdagger}}26.2 \\ 
Mixamo$\rightarrow$Kinetics \cite{da2022co2a} & \best{14} & \best{36,195} & 0.24 & 66.7 & \textsuperscript{\textdagger}\best{68.1} \\
\midrule
Kinetics$\rightarrow$BABEL & \second{12} & \second{18,946} & \best{0.31} & \best{182.1} & \hphantom{\textsuperscript{\textdagger}}\second{65.0} \\  
\bottomrule
\end{tblr}
}
\label{tab:dataset-comapare}

\raggedright
\footnotesize{$^*$The average number of videos across 6 settings.} \\
\footnotesize{\textsuperscript{\textdagger}The reported value from the paper.}
\end{table}


\section{Method}
\label{sec:method}
\begin{figure*}[ht]

    

\begin{center}

\mpage{.98}{
    \includegraphics[page=1,width=\linewidth]{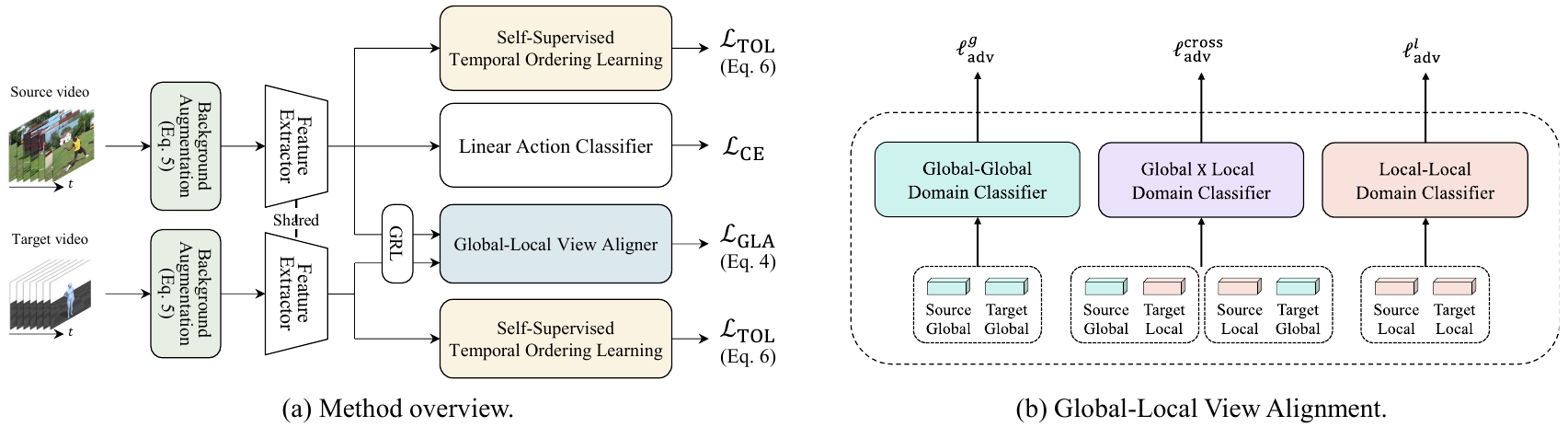}
}
\end{center}

\vspace{\figcapmargin}
\figcaption{Overview of GLAD}{
(a) GLAD consists of several key components. Firstly, we mix a video with a different background from another video to mitigate background bias. Next, a feature extractor extracts spatio-temporal feature vectors from the augmented videos. Then we feed the source feature vectors into a linear classifier to learn action labels. We employ a global-local view alignment module following a gradient reversal layer to align the source and target features. To further address background bias, the model learns the temporal order of shuffled clips in a self-supervised manner.
(b) To tackle the temporal shift between the source and target domains, GLAD utilizes three temporal view alignment methods: global-global, local-local, and global $\times$ local. Each method employs dedicated domain classifiers to align the source and target features.
}
\label{fig:overview}
\vspace{\figcapmargin}

\end{figure*}

We formulate the video action recognition task as an unsupervised video domain adaptation (UVDA). In UVDA, we have a labeled source video dataset $\mathcal{D}^s = \{ (x^s_i, y^s_i) \}$, where $x_i^s$ represents the input video and $y_i^s$ denotes the corresponding label, as well as an unlabeled target video dataset $\mathcal{D}^t = \{ x^t_i \}$. The source and target datasets share the same label space $\mathcal{K}$ between the source and target data. Our objective is to learn a model that performs well in the target domain. Simply applying a model trained solely on the source data to the target data leads to suboptimal performance~\cite{Ganin-ICML-2015,dann}. Therefore, a UVDA method should effectively leverage not only the labeled source data but also the unlabeled target data to achieve superior performance in the target domain.

We show an overview of the proposed method, GLAD, in \figref{overview} (a). Given a video, we mix it with a different background from another video for background debiasing (\secref{debiasing}). Then we extract a spatio-temporal feature vector from the augmented background mixed. We feed a source video feature vector into a linear classifier to learn actions with the standard cross-entropy loss. To align the source and target domains, we feed both the source and target feature vectors into the global-local view alignment module following a gradient reversal layer~\cite{Ganin-ICML-2015,dann} (\secref{gla}). To further mitigate the background shift between domains, we encourage the model to learn the temporal order of multiple clips in either a source or target video (\secref{debiasing}). We provide more details on each component in the following subsections.


\subsection{Global-Local View Alignment}
\label{sec:gla}
We propose Global-Local view Alignment (GLA) to align features of different domains even if action durations are significantly different across domains. As illustrated in \figref{teaser} (c), we observe action duration shifts across different domains, such as in the {\kb} dataset. For example, the \texttt{jump} action in Kinetics spans a duration of 10 seconds, involving a sequence of a run-up, a jump, and a landing. In contrast, the \texttt{jump} action in BABEL lasts only 1 second, consisting of a brief jump. Due to these temporal shifts, simply aligning the source and target feature vectors of clips using the same sampling strategy across domains may lead to suboptimal performance in UVDA, particularly when a large temporal distribution shift exists, as in the case of the {\kb} dataset.

\vspace{\paramargin}

\paragraph{Global and local temporal views.}
We define a uniformly sampled clip as a global clip and a densely sampled clip 
as a local clip.
For uniform sampling, we divide a video into equal-sized subsequences and randomly select one frame from each subsequence to construct a clip.
On the other hand, for dense sampling, we select frames with regular intervals, starting from a randomly chosen point, to construct a clip.

Let $\phi\noteg_m, \phi\notel_n$ denote global and local clip feature vectors, respectively, extracted by the feature extractor.
Then, we can define aggregated global/local feature vectors $\psi$ as follows:
\begin{equation}
    \psi\noteg = \frac{1}{M}\sum_{m=1}^M \phi\noteg_{m} \,\,, \quad \psi\notel = \frac{1}{N}\sum_{n=1}^N \phi\notel_{n} \,,
\end{equation}
where $M, N$ are the number of global and local clips sampled from a single video respectively.

\vspace{\paramargin}

\paragraph{Domain alignment.}
As shown in \figref{overview} (b), we employ individual domain classifiers to align feature vectors with different temporal granularities from the source and target domains. Specifically, we align the global feature vectors from the source and target domains (global-global), the local feature vectors from the source and target domains (local-local), and a global feature vector from one domain with a local feature vector from another domain (global-local). For the global-global alignment, we use an MLP denoted as $\mathcal{F}\noteg$. Similarly, we employ another MLP $\mathcal{F}\notel$ for local-local alignment and yet another MLP $\mathcal{F}\notex$ for cross-scale (global-local) alignment. To introduce adversarial training, we insert a gradient reversal layer (GRL)~\cite{Ganin-ICML-2015} between the feature extractor and the domain classifiers. The GRL negates gradients during backpropagation, effectively making the domain classifier adversarial.

Then, we define an adversarial loss $\ell_\adv$ for an arbitrary temporal view as follows:
\begin{equation}
    \ell_{\adv}(\mathcal{F}, \psi) \! =
        - \frac{1}{2B} \!\! \left[
        \sum_{i \le B} \log \mathcal{F}(\psi_i)
        \!+\! \sum_{i > B} \log (1 - \mathcal{F}(\psi_i)) \! \right] \!\!.
\end{equation}
Here, $B$ represents the batch size, and we differentiate between the two domains using their batch indices: $1 \le i \le B$ for the source domain and $B < i \le 2B$ for the target domain.
We define the final global-local view alignment loss as follows:
\begin{align}
    \mathcal{L}_{GLA}^{\vphantom{\noteg}} =
        \ell_{\adv}(\mathcal{F}\noteg, \psi\noteg)
        + \ell_{\adv}(\mathcal{F}\notel, \psi\notel)
        + \ell_{\adv}(\mathcal{F}\notex, \psi\notex),
        \label{eq:adv}
\end{align}
where $\psi\noteg, \psi\notel$, and $\psi\notex = (\psi\noteg, \psi\notel)$ denote the feature vectors for global-global, local-local, and global-local alignments, respectively. With the loss function \eqnref{adv}, we effectively align the feature vectors from different domains with significantly different action durations.

\subsection{Background Debiasing}
\label{sec:debiasing}
As depicted in \figref{teaser}, the {\kb} dataset exhibits a significant and realistic background distribution shift. To effectively address this background distribution shift, we incorporate two essential debiasing methods: i) background augmentation and ii) temporal order learning. These debiasing methods play a crucial role in enhancing the performance of UVDA, as demonstrated in the ablation study presented in \secref{ablation}. It is worth emphasizing that the careful selection and utilization of these debiasing methods contribute to achieving superior performance in UVDA tasks.


\vspace{\paramargin}

\paragraph{Background augmentation.}
To encourage a model to learn background-invariant representations, we employ a background augmentation technique. For each video in the dataset, we extract a background frame $b$ using a temporal median filter (TMF) \cite{piccardi2004tmf} and store these background frames for later use. The backgrounds obtained through TMF typically exhibit clear and appropriate backgrounds for the majority of videos.

During training, we randomly select a background from the stored background database and mix it with each frame of every video in a minibatch. We define the mixing process as follows:
\begin{equation}
        \tilde{x} (t) = (1 - \lambda) x (t) + \lambda b, \quad t = 1, \dots, T.
\end{equation}  
Here, $x(t)$ represents the $t$-th frame of the input video $x \in \mathbb{R}^{T \times H \times W \times C}$, $\lambda$ is a mix-up ratio uniformly sampled from the range $[0, 1]$. By providing action sequences against diverse backgrounds, we encourage the model to focus on the actions themselves rather than being overly influenced by the background context. This facilitates the learning of background-invariant representations that are essential for domain-adaptive action recognition~\cite{choi2020sava,sahoo2021comix}.

\vspace{\paramargin}

\paragraph{Temporal ordering learning.}
To account for significant background shifts across different datasets~\cite{choi2020sava}, we incorporate an additional learning objective, namely temporal order learning, to further regularize the model training in conjunction with background augmentation. We adopt the temporal clip order prediction~\cite{choi2020sava,xu2019cliporder} as a pre-text task for this purpose. 

In the clip order prediction task, the model tries to solve a puzzle of predicting the true order of $N$ shuffled clips. By solving this clip order prediction task, the model is encouraged to focus more on the action itself rather than being influenced by the static background. As illustrated in \figref{overview} (a), we feed both the source and target videos into the temporal order learning (TOL) module.

The TOL module shuffles the order of $N$ clip features $\boldsymbol{\phi} = \left( \phi_{n} \right)_{n=1}^N$ for each video. Consequently, we obtain $\widetilde{\boldsymbol{\phi}} = \left( \phi_{\sigma(n)} \right)_{n=1}^N$, where $\sigma$ denotes a permutation randomly chosen from the set of all possible permutations $\mathcal{S}_N$. We pass the shuffled clip features $\widetilde{\boldsymbol{\phi}}$ through a simple MLP, denoted as $\mathcal{F}_\Omega$, followed by a softmax operation to predict the correct order $\hat{\omega}_i \in [0, 1]^{N!}$, where $\sum_j \hat{\omega}_{i,j} = 1$. We define the TOL loss as follows:

\begin{equation}
    \mathcal{L}_{\cop} = -\frac{1}{2B \cdot N!} \sum_{i=1}^{2B} \sum_{j=1}^{N!} \omega_{i, j} \log \hat{\omega}_{i, j} \,.  
\end{equation}
A background-biased model is likely to struggle in predicting the correct order of clips, as its focus remains on the static background. Conversely, a model that focues on the actions is more likely to predict the correct order. By incorporating the TOL loss, we encourage the model to learn background-invariant representations.

\begin{table*}[t]
  \centering
  \caption{\textbf{Ablation study}. To validate the effect of each component, we show experimental results on the Kinetics$\rightarrow$BABEL dataset.  We conduct all experiments using the TSM~\cite{lin2019tsm} backbone. We report the mean class accuracy (MCA) with the corresponding standard deviation. The best performance is in \best{bold} and the second best is \second{underscored}.}


\begin{tabular}{cc}
\begin{minipage}{.34\textwidth}
\centering
{\fontsize{9pt}{10pt}\selectfont (a) Effect of various temporal alignments. \\} 
\resizebox{\linewidth}{!}{
{\small
\begin{tabular}{ccc|cc}
    \toprule
    \multicolumn{3}{c}{Temporal Alignment Strategies} & \multicolumn{1}{c}{MCA}\\
    \cmidrule(lr){1-3} \cmidrule(lr){4-4}
    Global-Global & Local-Local & Cross & K$\rightarrow$B \\
    \midrule

    \multicolumn{3}{c|}{\textcolor{gray}{Source-only baseline}}  & 18.5 $\pm$ 1.5 \\  
    \midrule
    \checkmark &  &                     & 25.5 $\pm$ 4.9 \\  
    & \checkmark &                      & 27.6 $\pm$ 3.7 \\  
    &  &\checkmark                      & 26.7 $\pm$ 1.6 \\  
    \checkmark & \checkmark &           & \second{28.4} $\pm$ 6.1 \\  
    \checkmark & \checkmark &\checkmark & \best{29.6} $\pm$ 1.7 \\  
    \bottomrule
\end{tabular}
}
}
\end{minipage}%
&
\begin{minipage}{.19\textwidth}
\centering
{\fontsize{9pt}{10pt}\selectfont (b) Effect of different temporal views. \\} 
\resizebox{\linewidth}{!}{
{\small
\begin{tabular}{cc|c}
    \toprule
    \multicolumn{2}{c}{Temporal View} & \multicolumn{1}{c}{MCA}\\
    \cmidrule(lr){1-2} \cmidrule(lr){3-3}
    Global & Local & K$\rightarrow$B \\
    \midrule
    1 & 0 & 18.5 $\pm$ 1.5 \\  
    0 & 1 & 21.5 $\pm$ 5.1 \\  
    \midrule
    3 & 0 & 28.3 $\pm$ 1.8 \\  
    2 & 1 & \best{30.1} $\pm$ 4.4 \\  
    1 & 2 & \second{29.6} $\pm$ 1.7 \\  
    0 & 3 & 25.1 $\pm$ 2.4 \\  
    \bottomrule
\end{tabular}
}
}
\end{minipage}


\begin{minipage}{.21\textwidth}
\centering
{\fontsize{9pt}{10pt}\selectfont (c) Effect of various debiasing methods.\\}
\resizebox{\linewidth}{!}{
{\small
\begin{tabular}{ccc}
    \toprule
    \multicolumn{2}{c}{Debias} & MCA\\
    \cmidrule(lr){1-2} \cmidrule(lr){3-3}
    Bg. Aug. & TOL &\multicolumn{1}{|c}{K$\rightarrow$B} \\
    \midrule
               &           & \multicolumn{1}{|c}{26.9 $\pm$ 3.2} \\  
    \checkmark &           & \multicolumn{1}{|c}{29.6 $\pm$ 1.7} \\  
    & \checkmark           & \multicolumn{1}{|c}{\second{33.4} $\pm$ 1.7} \\  
    \checkmark &\checkmark & \multicolumn{1}{|c}{\best{37.7} $\pm$ 2.5} \\  
    \bottomrule
\end{tabular}
}
}
\end{minipage}
\begin{minipage}{.24\textwidth}
\centering
{\fontsize{9pt}{10pt}\selectfont (d) Effect of combining GLA and background debiasing.\\}
\resizebox{\linewidth}{!}{
{\small
\begin{tabular}{ccc}
    \toprule
    \multicolumn{2}{c}{Method} & MCA\\
    \cmidrule(lr){1-2} \cmidrule(lr){3-3}
    Debiasing & GLA &\multicolumn{1}{|c}{K$\rightarrow$B} \\
    \midrule
               &           & \multicolumn{1}{|c}{26.4 $\pm$ 2.4} \\  
    \checkmark &           & \multicolumn{1}{|c}{\second{36.7} $\pm$ 3.6} \\  
    &\checkmark            & \multicolumn{1}{|c}{26.9 $\pm$ 3.2} \\  
    \checkmark &\checkmark & \multicolumn{1}{|c}{\best{37.7} $\pm$ 2.5} \\  
    \bottomrule
\end{tabular}
}
}
\end{minipage}
\end{tabular}

\label{tab:abl}
\end{table*}

\subsection{Training}

We define the final optimization objective as follows:
\begin{gather}
    \mathcal{L} := \mathcal{L}_\text{CE}(\theta_f, \theta_c)
        + \mathcal{L}_\cop(\theta_f, \theta_\sigma)
        - \mathcal{L}_{GLA}(\theta_f, \theta_d), \nonumber \\
    \vphantom{x}  \nonumber \\
    (\theta_f^*, \theta_c^*, \theta_\sigma^*) = \argmin_{\theta_f, \theta_c, \theta_\sigma}\mathcal{L}(\theta_d^*), \,
    \theta_d^* = \argmax_{\theta_d}\mathcal{L}(\theta_f^*, \theta_c^*, \theta_\sigma^*),
    \label{eq:obj}
\end{gather}
where $\theta_f, \theta_c, \theta_\sigma$, and $\theta_d$ denote the parameters of the feature extractor, action classifier, an MLP of TOL, and domain classifiers of GLA, respectively. 


\subsection{Inference}
During the inference stage, we remove all auxiliary components, including TOL and GLA, and retain only the feature extractor and linear action classifier. We do not utilize background augmentation.

Given an input video during inference, we extract one global feature vector and two local feature vectors. These features capture both global and local temporal information. We then average these feature vectors to obtain a single consensus feature vector that effectively represents the entire video. Finally, we feed the consensus feature vector into a linear classifier to predict the corresponding action label.

\label{sec:results}
\section{Experiments}
We conduct all the experiments on the {\kb} dataset. We use mean-class accuracy as an evaluation metric.




\subsection{Implementation details}
\label{sec:details}

We implement the proposed method using PyTorch and the mmaction library~\cite{2020mmaction2}. We choose I3D~\cite{carreira2018i3d} as the feature extractor for benchmarking against state-of-the-art methods, and TSM~\cite{lin2019tsm} for conducting ablation studies. The feature extractors are initialized with Kinetics400 pre-trained weights.
In the GLA module, we employ a 4-layer MLP for each domain classifier. To stabilize the training process, we employ curriculum learning~\cite{bengio2009curriculum}. We first pre-train the model with $\mathcal{L}_\cop$  for 500 epochs using 3 local clips to warm up the model. Then, we train the model with the final training objective \eqnref{obj} for 50 epochs.
We use SGD as the optimizer with a momentum of 0.9, a weight decay of 1e-4, and an initial learning rate of 2e-3. The learning rate is reduced by a factor of 10 at the 5th and 10th epochs. During warm-up, the batch size is set to 384 per GPU, while during the main training, it is set to 24 per GPU for both the source and target domains.
Background augmentation is applied only to the source domain clips, with a probability of 25\% and a fixed $\lambda$ value of 0.75. To better capture the temporal context in videos, we adopt two different sampling strategies: uniform sampling for global clips and dense sampling for local clips, maintaining a frame interval of 2 in both domains.
All experiments are conducted using 8 NVIDIA RTX 3090 GPUs.

\subsection{Ablation Study}
\label{sec:ablation}
We conduct an extensive ablation study to verify the effectiveness of each component and show the results in \tabref{abl}.



\vspace{\paramargin}
\paragraph{Effect of various temporal alignment methods in GLA.}
In \tabref{abl} (a), we show experimental results demonstrating the impact of different temporal alignment methods in the GLA module. The \emph{Global-Global} refers to employing a domain classifier with global clips from source and target domains. \emph{Local-Local} refers to employing a domain classifier with local clips. \emph{Cross} refers to employing a domain classifier with a global clip from one domain and a local clip from another domain.
As shown in the table, incorporating both global and local alignments leads to superior performance (28.4\%) compared to focusing on either one alone (25.5\%, 27.6\%). Notably, we achieve the highest performance of 29.6\% when we employ all three alignments together. Furthermore, the collaborative operation of these alignment methods results in a relatively more stable performance, as indicated by the lower standard deviation value.


\vspace{\paramargin}
\paragraph{Effect of the number of global and local views.}
From the results presented in \tabref{abl} (b), aligning only local clips surpasses aligning only global clips (21.5\% versus 18.5\%). However, combining both global and local alignments leads to even higher accuracy. Specifically, employing a combination of two global and one local view per video for alignment achieves the highest accuracy of 30.1\% with a standard deviation of 4.4. Notably, using one global and two local views per video for alignment demonstrates comparable accuracy of 29.6\%, with a lower standard deviation of 1.7. Based on these findings, we utilize one global and two local views per video for alignment in the subsequent experiments.


\vspace{\paramargin}
\paragraph{Effect of background debiasing.} 
As shown in \tabref{abl} (c), both background augmentation and TOL demonstrate performance improvements of 2.7 and 6.5 points, respectively, compared to the baseline without debiasing. Furthermore, when we combine both debiasing methods, we observe a substantial gain of 10.8 points. These results highlight the complementary nature of the two debiasing methods, emphasizing the importance of employing them together. Please note that GLA is enabled for all experiments conducted.


\vspace{\paramargin}
\paragraph{Complementary nature of GLA and background debiasing.}
\tabref{abl} (d) demonstrates the complementary nature of background debiasing and GLA. When applying background debiasing without GLA, we observe a substantial improvement of 10.3 points compared to the baseline. Similarly, applying GLA without debiasing results in a modest improvement of 0.5 points. However, when we employ both debiasing and GLA together, we achieve a remarkable improvement of 11.3 points compared to the baseline, with a lower standard deviation (3.6 \vs 2.5). These results clearly indicate that the two methods are complementary to each other, generating a synergistic effect that enhances the overall performance of the UVDA model.


\subsection{Comparison with state-of-the-arts}
\begin{table}[t]
\centering
\caption{\textbf{Comparison with state-of-the-art on the Kinetics$\rightarrow$BABEL dataset.}
We show the mean class accuracy (MCA) \
For a fair comparison, we indicate 
the number of clips $N_c$ and the number of frames per clip $N_f$.
All methods employ I3D~\cite{carreira2018i3d} as the backbone. The best performance is in \best{bold} and the second best is \second{underscored}.}

\resizebox{0.9\columnwidth}{!}{
\begin{tabular}{ l  c  c  }
      \toprule
Method & $N_c \times N_f$ & Kinetics$\rightarrow$BABEL  \\
\midrule
Source only  & $3 \times 8 = 24$ & 11.7 $\pm$ 0.7 \\  %
DANN ~\cite{Ganin-ICML-2015} & $3 \times 8 = 24$  & \second{29.3} $\pm$ 1.5  \\
CoMix ~\cite{sahoo2021comix} & $16 \times 8 = 128$ & 21.4 $\pm$ 0.3  \\
CO2A ~\cite{da2022co2a} & $4 \times 16 = 64$ & 24.7 $\pm$ 0.8   \\
\textbf{GLAD (Ours)} & $3 \times 8 = 24 $ & \best{33.7} $\pm$ 1.8  \\  
\midrule
Supervised target& $3 \times 8 = 24$ & 76.7 $\pm$ 2.1  \\  
      \bottomrule
\end{tabular}
}
\label{tab:main_kinetics-babel}
\end{table}


\label{sec:sota}
In this section, we compare the proposed method with state-of-the-art UVDA methods. We show the results in \tabref{main_kinetics-babel}. ``Source only'' refers to the baseline method of training on 
labeled source data and testing on target data, which sets the lower bound for UVDA.
``Supervised target'' is an upper bound performance: a model trained with target data with labels. DANN~\cite{Ganin-ICML-2015} is an image-based domain adaptation method extended to UVDA. CoMix~\cite{sahoo2021comix} and CO2A~\cite{da2022co2a} are state-of-the-art UVDA methods.
Surprisingly, we observe that the simple DANN method outperforms CoMix and CO2A on the challenging {\kb} dataset. Our proposed method, GLAD, achieves the highest performance of 33.7\%, surpassing DANN by 4.4 points. Notably, we achieve superior results with significantly fewer clips and frames compared to CoMix and CO2A, which highlights the high efficiency and accuracy of the proposed method.

\subsection{Qualitative evaluation}

\begin{figure}[t]
\centering
\includegraphics[page=1,width=\linewidth]{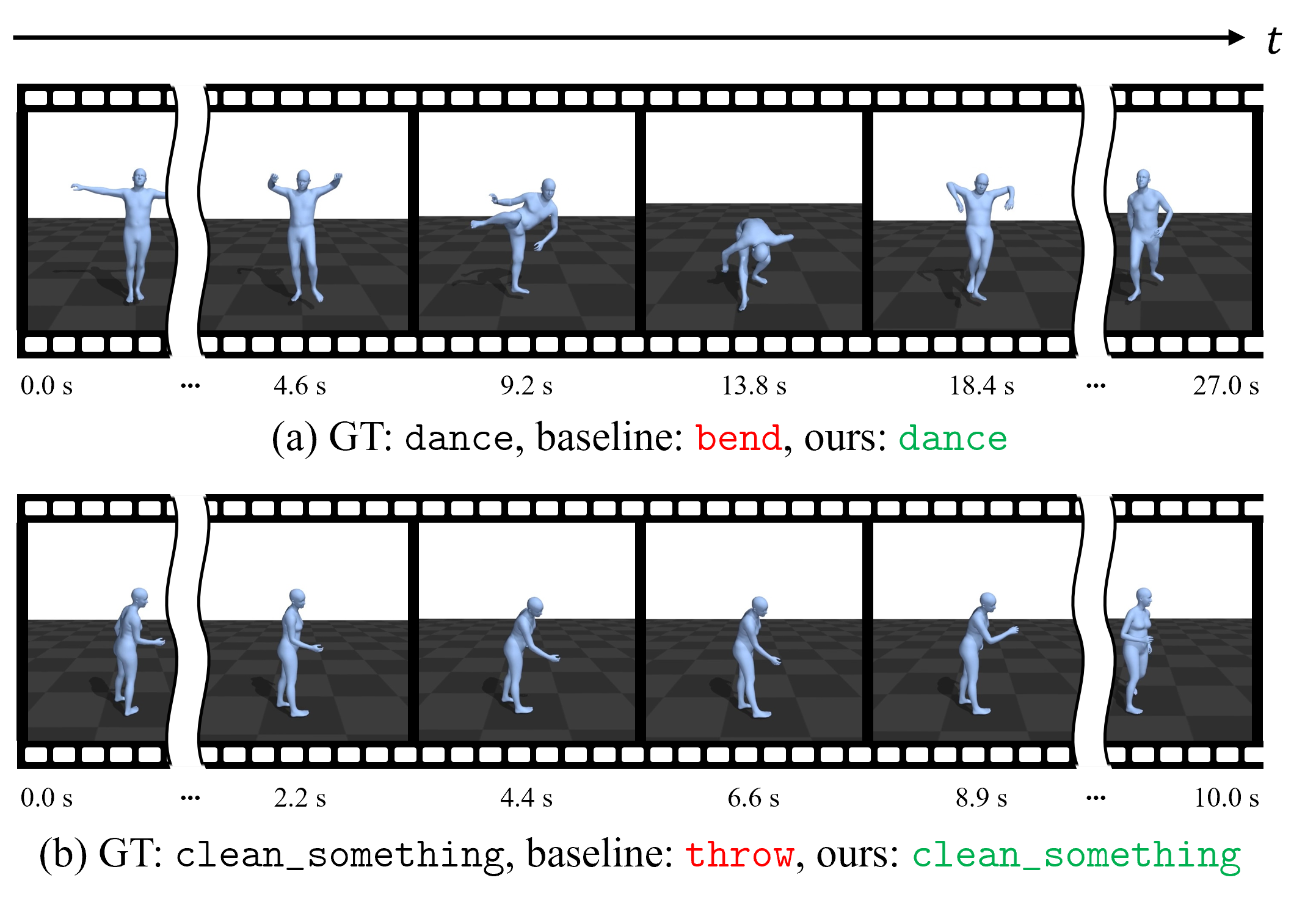}
\caption{\textbf{Qualitative examples from Kinetics$\rightarrow$BABEL.} We compare predictions of ours (GLAD) with predictions of a baseline (DANN \cite{dann}). GT denotes ground-truth, correct predictions are in \textcolor[HTML]{00B050}{green}, and incorrect predictions are in {\color{red}red}. We observe the baseline fails to predict correct actions due to the challenging temporal gap while GLAD consistently predicts correct actions.}
\label{fig:qual}
\end{figure}

In \figref{qual}, we show some qualitative examples from the Kinetics$\rightarrow$BABEL to validate the effectiveness of GLAD. We compare the predictions of the baseline (DANN \cite{dann}) and GLAD on the BABEL dataset. The ground-truths for the four example videos are \texttt{dance}, \texttt{clean\_something}, \texttt{crawl} and \texttt{pick\_up} with durations of 27.0, 10.0, 2.7 and 1.9 seconds, respectively.
In the example shown in \figref{qual} (a) with \texttt{dance} action, the baseline fails to understand a long video of 27.0 seconds. The prediction \texttt{bend} implies the model focuses only on the bending motion which lasts for only 3 seconds in the video. The result might imply that the baseline tries to focus on a few key frames or local motions instead of focusing on the global temporal context when the action duration differs from the source data.
Furthermore, for the example shown in \figref{qual} (b), the baseline fails to distinguish \texttt{clean\_something} from \texttt{throw} which involves understanding different speeds. In contrast, GLAD correctly predicts \texttt{clean\_something}.






\section{Conclusions}
\label{sec:conclusions}


In this paper, we have addressed the challenging problem of unsupervised video domain adaptation for action recognition, specifically focusing on scenarios with a significant domain gap between the source and target domains. To overcome the limitations of existing datasets that are small in scale and lack significant domain gaps, we have introduced the {\kb} dataset, which provides a more challenging, realistic, and large-scale benchmark. Our proposed method, GLAD, incorporates global-local view alignment to tackle temporal distribution shifts and background debiasing to address background distribution shifts.
We have demonstrated the effectiveness of our proposed method through extensive experiments. 
Despite using fewer clips and frames compared to existing methods, GLAD has achieved favorable performance. 
The promising results highlight the efficacy and efficiency of our proposed method, paving the way for further advancements in unsupervised video domain adaptation for action recognition. 




\vspace{-1em}
\paragraph{Acknowledgment.}
This work is supported 
by NCSOFT; 
by the Institute of Information \& Communications Technology Planning \&
Evaluation (IITP) grant funded by the Korea Government (MSIT) (Artificial Intelligence Innovation Hub) under Grant 2021-0-02068; 
by the National Research Foundation of Korea(NRF) grant funded
by the Korea government(MSIT) (No. 2022R1F1A1070997).

{\small
\bibliographystyle{setup/wacv/ieee_fullname}
\bibliography{main}
}

\end{document}